\documentclass[conference]{IEEEtran}
\IEEEoverridecommandlockouts

\usepackage{amsmath}
\usepackage{amssymb}
\usepackage{amsfonts}
\usepackage{booktabs}
\usepackage{graphicx}
\usepackage{multirow}
\usepackage{array}
\usepackage{makecell}
\usepackage{braket}
\usepackage{textcomp}
\usepackage{xcolor}
\usepackage{soul}
\usepackage{bbding}
\usepackage{pifont}
\usepackage{wasysym}

\usepackage[numbers]{natbib}
\usepackage{notoccite}
\usepackage{url}

\usepackage[T1]{fontenc}
\usepackage[utf8]{inputenc}

\usepackage{float}

\usepackage[bookmarks=true]{hyperref}

  \graphicspath{{fig/}}
  \DeclareGraphicsExtensions{.pdf,.jpeg,.png}

\begin{document}
\title{\vspace{4mm}%
A multi-modal tactile fingertip design for robotic hands to enhance dexterous manipulation}

\author{Zhuowei~Xu$^{*}$,
        Zilin~Si$^{*}$,
        Oliver~Kroemer,
        and~Zeynep~Temel%
\thanks{$^{*}$Equal contribution.}%
\thanks{All authors are with Carnegie Mellon University.}%
}

\maketitle

\begin{abstract}
Tactile sensing is critical for enhancing manipulation precision and versatility, but its adoption in robotic hands remains limited due to high sensor costs, manufacturing and integration challenges, and difficulties in extracting expressive and reliable contact information.
In this work, we present a low-cost, easy-to-make, adaptable, and compact fingertip design for robotic hands that integrates multi-modal tactile sensors. We use strain gauge sensors to capture static forces and a contact 
microphone sensor to measure high-frequency vibrations during contact. These tactile sensors are integrated into a compact design with a minimal sensor footprint. From sensor characterization, we show that strain gauge sensors provide repeatable 3D force measurements in the 0-5~N range and the contact microphone sensor can captures distinct vibrotactile signatures associated with different material types. We evaluate our design with three types of manipulation tasks, including slip-aware closed-loop grasping to handle a wide range of objects, cup counting and unstacking with precise contact localization, and box selection and opening with hidden material identification. Given the expressiveness and reliability of tactile sensor readings, we show that different tactile sensing modalities can be used flexibly in manipulation, solely or together to achieve improved task performance. The design will be open-sourced on the project website: \url{https://sites.google.com/view/tactilefingertip}.

\end{abstract}


\section{Introduction}

\begin{figure}[!t]
\centering
\includegraphics[width=0.90\linewidth]{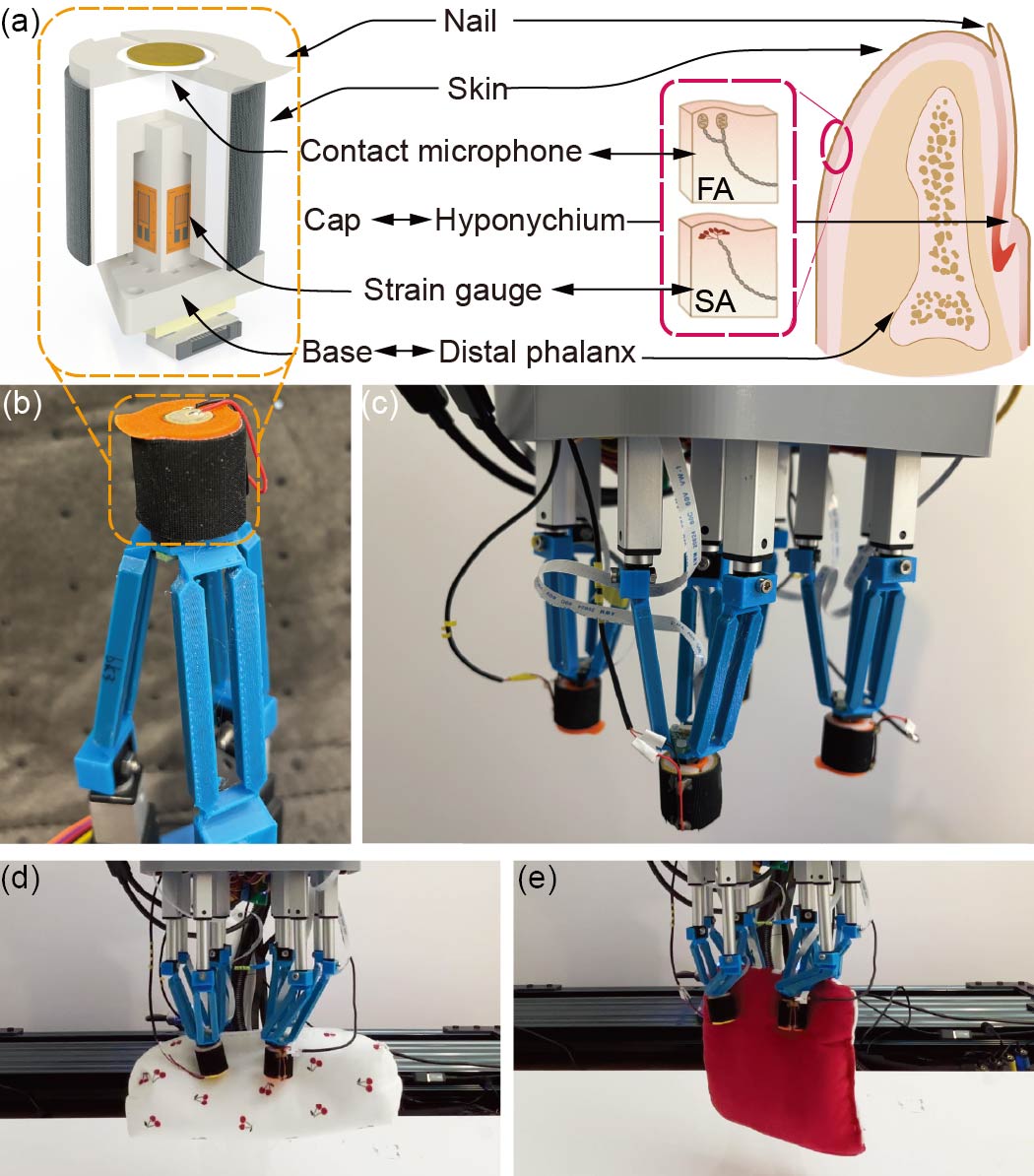}
\caption{(a) Human-inspired multi-modal tactile fingertip design: the strain gauge sensors emulate slow-adapting (SA) mechanoreceptors for pressure sensing, while the contact microphone emulates fast-adapting (FA) mechanoreceptors for sensing contact vibrations and dynamic events such as slip~\cite{kandel2000principles}. (b) A Delta finger~\cite{10380685} equipped with the proposed fingertip. (c) A DeltaHand~\cite{10380685}  equipped with four tactile fingertips. (d) (e) Adaptively grasp a soft bag and a sandbag by using multi-modal tactile sensing in control loop to detect contact and slip.}
\label{fig:teaser}
\vspace{-6mm}
\end{figure} 


Humans rely on tactile feedback from diverse mechanoreceptors in the hand to perceive pressure, detect slip, and infer properties of contact surfaces~\cite{bullock2011classifying,papagno2016deaf}. These sensory cues are seamlessly integrated to enable dexterous and precise manipulation. Towards contact-rich and dexterous manipulation, multi-fingered robotic hands have been developed to emulate the functionality of the human hand, and the integration of multi-modal tactile sensing in robotic hands has become increasingly important for handling complex contact scenarios involving multiple simultaneous interactions and extended object coverage~\cite{kappassov2015tactile}.

Despite the wide range of tactile sensing technologies in the literature ~\cite{svensson2021electrotactile, mejia2024hearing,kim2022soft, chen2023intrinsic,alfadhel2016magnetic, yan2021soft, dai2022design, shang2026forte}, developing and deploying tactile sensors for robotic hands presents unique challenges. Robotic hands demand tactile sensors that are compact enough to integrate into fingers without restricting hand's workspace or mobility. The nature of contact-rich manipulation, where dexterous hands repeatedly make, break, and switch contacts, requires stable and reliable sensor measurements without the sensor's tear and wear. Inspired by the richness of human touch, multi-modal tactile sensing can expand the hand's versatility by capturing nuanced contact information such as forces, slippage, and textures. In addition, expressive and interpretable sensor signals can be effectively and seamlessly incorporated into the high-DoF robotic hand control systems.

\begin{table*}[t]
  \centering
  \small
  \renewcommand{\arraystretch}{1.2}
  \begin{tabular}{p{2cm}ccccccc}
    \hline
    {\textbf{Tactile sensor}}  & \multicolumn{2}{c}{\textbf{Dynamic sensing}} & \multicolumn{2}{c}{\textbf{Static sensing}} & \textbf{Size (mm)} & \textbf{Open-source} & \textbf{Cost (USD)}\\
    \cmidrule(lr){2-3}\cmidrule(lr){4-5}
    \textbf{} & \textbf{Modality} & \textbf{Frequency} & \textbf{Modality} & \textbf{Frequency} & \textbf{} & \textbf{} & \textbf{} \\
    \hline
    GelSight~\cite{yuan2017gelsight} & - & - & Visual & 25 Hz & $32 \times 28 \times 28 $ & \XSolidBrush & 510 \\
    CoinFT~\cite{choi2025coinft} & - & - & Capasitive & 360 Hz  & $20 \times 20 \times 2 $ & \CheckmarkBold & 11 \\
    Digit360~\cite{lambeta2024digit360} & Audio & 10 kHz & Visual \& Pressure & 240 Hz & $25 \times 25 \times 26 $ & \CheckmarkBold & - \\
    SpikeATac~\cite{chang2025spikeatac} & PDVF & 4 kHz & Capacitive  & 40 Hz & $45 \times 32 \times 25 $& \XSolidBrush & 365 \\
    {\textbf{Ours}} & Audio & 44.1 kHz & Resistive & 90 Hz & $19 \times 19 \times 27 $  & \CheckmarkBold & $ 90 $ \\ 
    \hline
  \end{tabular}
  \caption{Tactile sensors for robotic hands in literature. Our design includes both dynamic and static sensing modality, and it is also comparably small in size and low in cost with open-sourced design for easy adaptation to other robotic end-effector.}
  \label{tab:sensor-compare}
  \vspace{-6mm}
\end{table*}

In this work, we present a multi-modal tactile fingertip that integrates complementary sensing in a compact design (Fig.~\ref{fig:teaser}(a)). We embed strain gauge sensors and a contact microphone sensor in the fingertip structure to capture force and vibration generated through contact, reflecting the slow and fast mechanoreceptors of human touch. These sensors are internal to the fingertip and therefore avoid direct wear and tear from interactions. We use off-the-shelf sensors to ensure accessibility, facilitate replication across platforms, and reduce fabrication overhead with minimal sensor reconfiguration when adapting the design to different robot end-effectors. Our design is compact (\(19\,\mathrm{mm}\times19\,\mathrm{mm}\times27\,\mathrm{mm}\)), cost-effective (\$90), and easy to fabricate (within an hour). Through sensor characterization, we demonstrate that the tactile fingertip can reliably predict 3D force and distinguish contact surface properties. By equipping four tactile fingertips on a DeltaHand~\cite{10380685} (Fig.~\ref{fig:teaser}(c)), we demonstrate the ease of integration with multi-finger robotic hands and their applications in robotic manipulation. 


The main contributions of this work are threefold: (i) a multi-modal tactile fingertip design that integrates force and vibration sensing in a compact form factor suitable for multi-finger robotic hands; (ii) sensor characterization showing that the fingertip provides reliable and expressive tactile measurements; and (iii) evaluation on three representative manipulation tasks: closed-loop grasping, cup counting and unstacking, and hidden object infer to show how multiple tactile modalities can be flexibly used or fused to handle fragile, cluttered, unknown objects. Across these tasks, the results show that multi-modal tactile sensing supports robust grasping of challenging objects such as transparent or deformable objects, and improves performance in contact-rich and dexterous manipulation under visual occlusion.


\begin{figure*}[!h]
\centering
\includegraphics[width=1\linewidth]{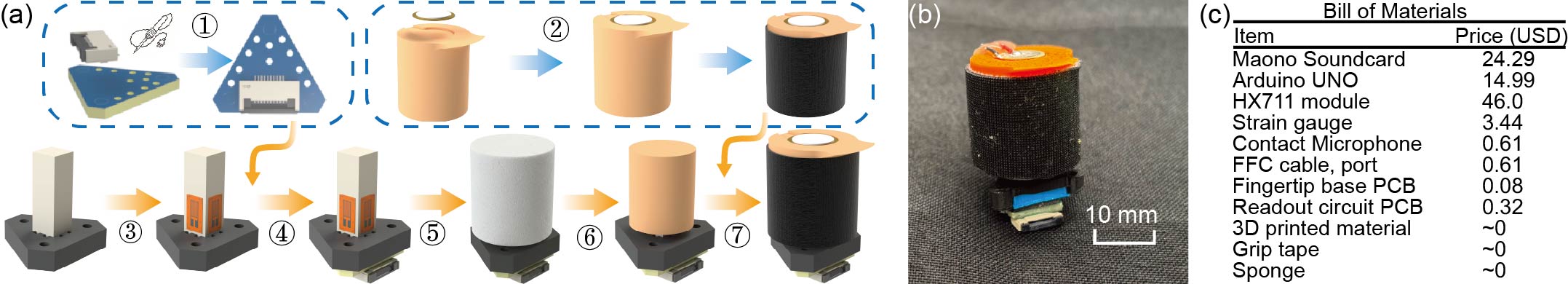}
\caption{The fabrication procedure and final prototype of the fingertip. (a) Fingertip fabrication: $\textcircled{1}$ Soldering the connector onto the PCB; $\textcircled{2}$ attaching the contact microphone sensor on the top of the 3D-printed cap using super glue; $\textcircled{3}$ gluing strain gauge sensors on four face of the square prism which is 3D-printed together with the fingertip base; $\textcircled{4}$ mounting the PCB beneath the base with M2 screws; $\textcircled{5}$ gluing the 3D-printed cap on top of the base; $\textcircled{6}$ applying a layer of foam on the cap; $\textcircled{7}$ attaching grip tape as finger skin. (b) A fully assembled fingertip prototype. (c) The bill of materials for one fingertip.}
\label{fig:design}
\vspace{-3mm}
\end{figure*}

\section{Related Work}


\subsection{Multi-modal tactile sensor design}
To acquire complementary tactile information, researchers have developed multi-modal sensors that integrate different sensing modalities. For example, BioTac~\cite{fishel2012sensing} employs fluid conduction to sense force, vibration, and temperature, while vision-based sensors such as GelSight~\cite{yuan2017gelsight} capture high-resolution contact geometries and force distributions. More recently, Digit 360~\cite{lambeta2024digit360} combines optical, pressure, vibrotactile, and temperature sensing in a compact form. Despite their capabilities, vision-based sensors are often bulky and susceptible to skin wear, while fluid-based sensors tend to be costly and require complex manufacturing and signal processing, limiting their integration into robotic hands. In contrast, single-modal force sensors~\cite{jung2015piezoresistive, koiva2013highly, el2024compact} and vibration sensors~\cite{dai2022design, yang2023touch, chang2024investigation} have been widely adopted for their robustness, sensitivity, and compactness. To leverage the benefits of both modalities, we introduce a simple yet effective multi-modal fingertip that integrates off-the-shelf strain gauge sensors for force sensing with a contact microphone sensor for vibration sensing. This design offers complementary tactile signals for contact-rich manipulation, while remaining low-cost, compact, durable, and easy to integrate, requiring only a one-time calibration. Quantitative comparison with other sensors can be seen in Table~\ref{tab:sensor-compare}.

\subsection{Manipulation with tactile sensing}
Tactile sensing plays a dual role in robotic manipulation, serving both as direct feedback for control and as a source of perceptual features for high-level inference~\cite{li2020review}.
At the control level, force feedback has been used to guide fingers toward feasible grasp locations~\cite{fearing2003simplified}, and modern approaches often integrate force/torque control with optimization-based motion planning~\cite{righetti2014autonomous}. Tactile and force signals have also supported fine-grained control, such as detecting contact phase transitions for smooth grasping~\cite{howe1990grasping}, estimating and regulating grip force through slip detection~\cite{tremblay1993estimating}, and localizing and manipulating objects in-hand using pressure sensing arrays~\cite{chebotar2014learning}. For robot perception, tactile information can be used to classify object materials~\cite{dai2022design}, or guide robots along restricted structures such as cable routing~\cite{wilson2023cable}. Building on these insights, we investigate how multi-modal tactile sensing can be directly used to inform hand control or enable tactile perception through active interactions and then guide subsequent manipulation.

\section{Methodology}

We present our fingertip design (Fig.~\ref{fig:teaser}(a)), which integrates functional structures (a rigid fingernail and a compliant friction layer) and two tactile sensors (strain gauge sensors and a contact microphone), into a compact fingertip module. These sensors provide complementary tactile information, are easy to integrate due to their flat and compact form factors, and produce interpretable, reliable measurements for robotic hand control and contact-rich manipulation.

\subsection{Fingertip design and fabrication}

As shown in Fig.~\ref{fig:design}, the fingertip comprises five main components: 1) a triangular base, 2) a compliant square prism, 3) a rigid inner cap, 4) a foam isolation layer, and 5) a rigid outer cap with an integrated fingernail and grip tape covering. The triangular base serves as the mounting interface for the robotic finger and the custom printed circuit board (PCB) used for sensor data acquisition. Strain gauge sensors are attached to the four faces of the compliant square prism where the deformation can be captured as tactile signals. A rigid inner cap encloses the prism and provides structural support for the overlying foam layer. The foam layer is used to isolate the contact microphone from the noise of the robot conducted through the fingertip base. The rigid outer cap with a \(3\,\mathrm{mm}\)-extended fingernail encloses the foam layer and houses the contact microphone at the top. Finally, a soft grip tape layer covers the outer cap surface to increase friction and improve grasp stability. The overall fingertip has compact dimensions of \(19\,\mathrm{mm}\times19\,\mathrm{mm}\times27\,\mathrm{mm}\).

Fig.~\ref{fig:design}(a) illustrates the fabrication procedure. The triangular base and compliant prism are 3D-printed together using a Bambu Lab printer, with PETG for the base and TPU for the prism, at  90\% infill. The inner and outer caps are separately 3D-printed in PLA. After four strain gauge sensors are super-glued to the prism, the inner cap is super-glued onto the prism and the PCB is mounted beneath the base using screws. A \(3\,\mathrm{mm}\)-thick foam layer is taped to the inner cap, and then the outer cap is taped to enclose it. Finally, grip tape is wrapped around the outer cap and a contact microphone is super-glued on top. The full fabrication process takes approximately one hour including 3D printing. The total cost of materials and electronic components for one fingertip is \$90 and the bill of the materials is listed in Fig.~\ref{fig:design}(c).
Owing to its low cost and straightforward fabrication, the design is scalable for batch production and deployment on multi-finger robotic hands.

\subsection{Strain gauge sensor}
We use four strain gauge sensors (BF120-3AA, \(4\times6.8\,\mathrm{mm}\)) to estimate 3D spatial forces applied on the lateral and top surface of the fingertip. The strain gauges are compliant and deform with the surfaces they are bonded to. This deformation induces strain changes in the sensors, which are further transduced into electrical signals. The four strain gauges are arranged orthogonally to capture different components of the contact force.

\subsection{Contact microphone sensor}
To sense the dynamic contact events, we use a piezo disc contact microphone sensor (with $10$ mm diameter) to capture the vibrations generated through dynamic contact, such as making and breaking contact, and sliding. The sensor is rigidly attached on top of the fingertip cap, located close to the vibration source, i.e., the fingernail. When the fingernail interacts with an object, the contact vibration is transmitted through the rigid nail of the cap, captured by the sensor, and then transduced to electrical signals. 

\begin{figure}[!t]
\centering
\includegraphics[width=0.9\linewidth]{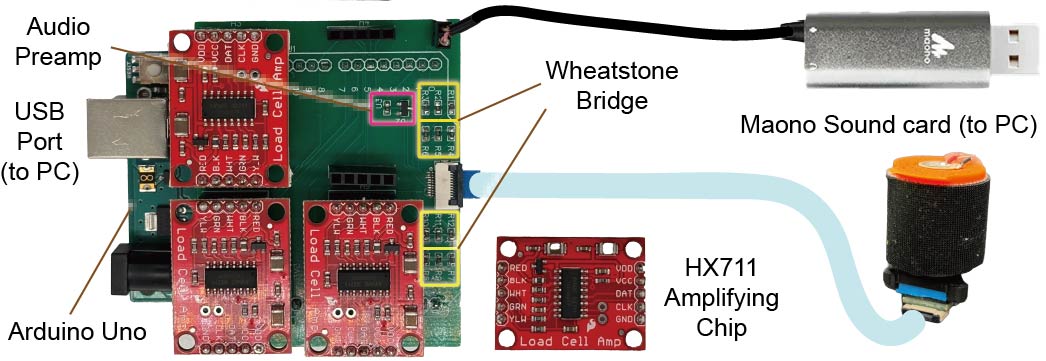}
\caption{Readout circuit of the fingertip: The contact microphone and the strain gauge sensors are read out through the fingertip PCB and routed via an FFC cable to an off-robot PCB. The strain gauge sensor measurements are amplified and digitized with an HX711 module on an Arduino Uno, while the vibrotactile signals are pre-amplified, digitized by a Maono USB sound card. Both are transmitted to the PC via USB.}
\label{fig:circuit}
\vspace{-3mm}
\end{figure}

\subsection{Read-out circuit}
\begin{figure*}
\centering
\includegraphics[width=1\linewidth]{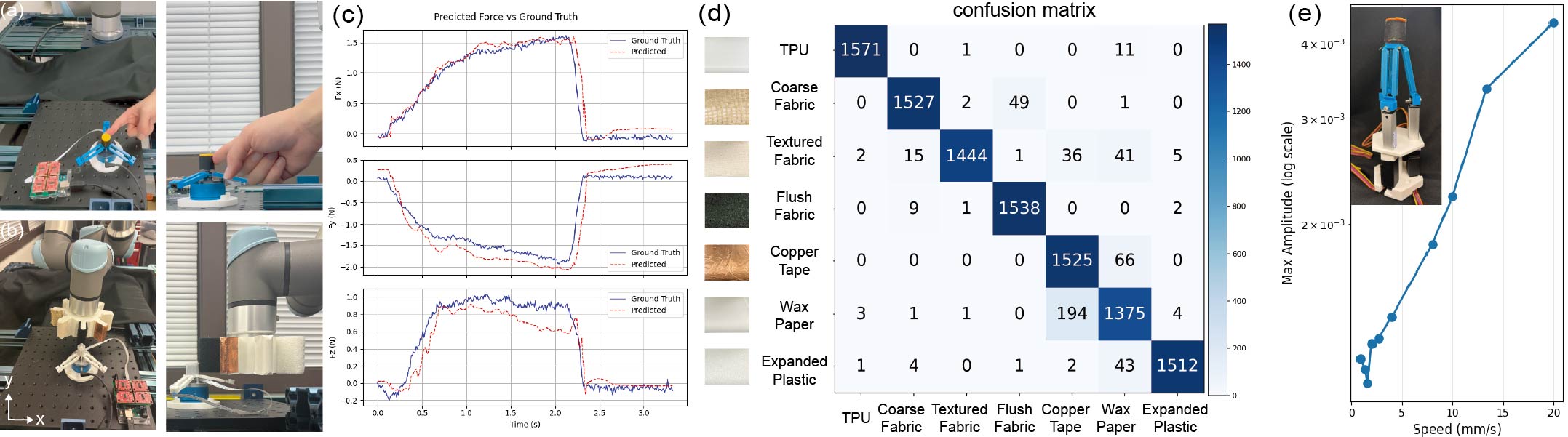}
\caption{Tactile sensor characterization setup and results: (a) The strain gauge sensor characterization setup. A human randomly interacts with the fingertip while recording the strain gauge sensor measurements and ground truth force readings from a force/torque sensor. (b) Contact microphone sensor characterization setup. We use a custom end-effector with samples of seven materials attached on different faces. The UR5e arm is controlled to initiate sliding contacts between the fingernail and a sample to generate vibrotactile signals. (c) Comparison between predicted and ground truth 3D forces. (d) Material classification error matrix based on vibrotactile measurements. (e) Contact microphone's sensitivity to the noise from Delta finger's linear actuator.}
\label{fig:setup}
\vspace{-3mm}
\end{figure*}

A customized PCB with the same footprint as the fingertip is used for sensor data acquisition. It takes signal inputs from both the strain gauge sensors and the contact microphone sensor and outputs them through a flat flexible cable (FFC) connector. The signals are then transmitted through aN FFC for clean and reliable cable routing. 

The FFC is then connected to another off-robot customized PCB (Fig.~\ref{fig:circuit}) for signal amplification and analog-to-digital conversion. The signals from the contact microphone are amplified with a preamp and digitized with an off-the-shelf sound card. Wheatstone bridges in standard configuration are used to measure the resistance changes of strain gauge sensors. The analog output of the bridge is fed into an HX711, which provides amplification and analog-to-digital conversion. The resulting digital signals are read out by an Arduino Uno and then transmitted to a computer via USB. The response frequencies of the strain gauge sensors and the contact microphone sensor are \(90\,\mathrm{Hz}\) and \(44.1\,\mathrm{kHz}\), respectively.

\section{Sensor Characterization Experiments}
To understand the sensor's capability and precision, we performed two sets of characterization: i) force estimation with strain gauge sensors and ii) material classification and noise characterization with the contact microphone sensor. 

\subsection{Force estimation with strain gauge sensors}
\label{sec:force-characterization}

We characterize the sensitivity and accuracy of the strain gauge sensors by mapping the raw sensor measurements to 3D forces with a multilayer perceptron (MLP) model. As shown in Fig.~\ref{fig:setup}(a), the fingertip is rigidly mounted on top of a 6-DOF force torque sensor (Nordbo NRS-50), which provides the ground-truth force measurements. We first collect data for model training, where a human operator uses their fingers to randomly apply different loads on the fingertips with different angles and magnitudes. The data collection lasts for 10~min and includes 54{,}000 data points.



The raw 4D input is expanded to a 16D feature vector by applying the element-wise transformations \([x, x^2, x^3, e^{-x}]\). We then trained a five-layer multilayer perceptron (MLP) model with a 16D input and a 3D output using the collected data. During online testing, the model achieves a mean absolute error (MAE) of \(0.07\,\mathrm{N}\) on each dimension in the range of \(0\text{--}5\,\mathrm{N}\). We illustrate the comparison between the predicted forces of the MLP model and the ground truth provided by the 6-DOF force torque sensor in Fig.~\ref{fig:setup}(c). 

\subsection{Material classification with the contact microphone sensor}

Contact microphone sensors can pick up vibration signals generated during dynamic interaction. We characterize the capability of the contact microphone sensor in classifying materials with sliding motions. 

We use seven materials including \textit{TPU}, \textit{Expanded Plastic}, \textit{Copper Tape}, \textit{Coarse Fabric}, \textit{Flush Fabric}, \textit{Wax Paper}, and \textit{Textured Fabric}. A customized end-effector is attached to a UR5e arm, with each material mounted on a different face. For each material, the UR5e arm rotates to align with the target face, approaches the fingertip to initiate contact, and then slides vertically for \(16\,\mathrm{mm}\) while recording vibrotactile signals. In total, 20~min of samples are collected for all materials. The raw time-domain signals are first transformed into frequency-domain spectrograms using Short-Time Fourier Transform (STFT), then cut into chunks by a window with a length of 5 frames (\(0.07\,\mathrm{s}\)). Finally, we train a five-layer multilayer perceptron (MLP) classifier that takes 10{,}240D spectrograms as input and outputs a 7D one-hot encoded vector representing the seven material classes.

The classification model achieves an overall 95.49\% accuracy on the test set, and we show the confusion matrix in Fig.~\ref{fig:setup}(d). Most misclassifications occurred between \textit{Copper Tape} and \textit{Wax Paper}. We attribute this confusion to the fact that both materials have very smooth surfaces, which produce similar vibrotactile signals during interaction.

To investigate the contact microphone's sensitivity to the environmental noise, we quantify the effect of the actuator-induced noise on the contact microphone because the hand's actuators are the closest and loudest noise resource. We use one Delta finger with a fingertip, and command it to extend $40\,\mathrm{mm}$ at speeds ranging from $1$ to $20\,\mathrm{mm/s}$ while recording the contact microphone readings. Without any contact, the readings represent the noise generated from the actuators. Fig.~\ref{fig:setup}(d) shows that the maximum noise amplitude increases with speed but remains at the level of $10^{-3}$, with a worst-case peak of $4.3 \times 10^{-3}$. This indicates that actuator-induced noise always stays below the signal level later used to detect slip events which is $6.0 \times 10^{-3}$.

\section{Experiments}
\begin{figure*}[ht]
    \centering
    \includegraphics[width=0.99\linewidth]{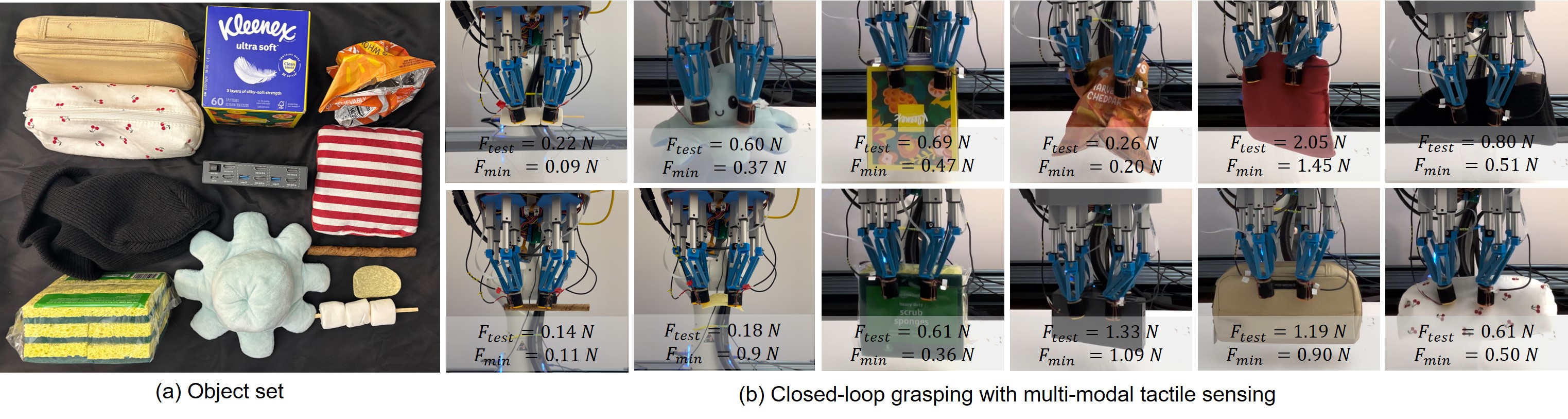}
    \caption{(a) Object set for adaptive grasping. We select objects with diverse grasping challenges, including deformable, fragile, and irregular objects. (b) We demonstrate successful grasping by using force control to initiate the grasp and vibrotactile signals to detect slip and adjust grasp strength. The labeled force values include $F_{\text{test}}$, the average normal force per fingertip after lifting with our grasping strategy, and $F_{\min}$, the minimum grasping force required to prevent dropping, obtained by stepwise release until failure.}
    \label{fig:pinching-task}
\vspace{-3mm}
\end{figure*}
\subsection{Overview}
We apply the proposed tactile fingertips on three manipulation tasks: \\
\textbf{Adaptive grasping with slip detection}: we develop a closed-loop hand control with force control and vibrotactile-based slip detection to grasp challenging objects which are fragile, deformable, or with irregular shapes with minimal force applied to avoid damage; \\
\textbf{Cup localization and unstacking}: we use vibrotactile sensing with sliding motions to locate the cup edges and guide to pick up the top cup from a stack. This task involves fine-grained features that are challenging to perceive with vision and to manipulate without using fingernail; \\
\textbf{Hidden object perception through active interaction to guide manipulation}: Given visually occluded boxes, we explore and identify hidden material properties through fused force and vibrotactile sensing during active lifting and shaking motions and use it to inform subsequence manipulation, such as choosing the desired box to open.

Through these tasks, we aim to show the functionality of the designed fingertips and how multi-modal tactile sensing can be used in closed perception-manipulation loops.


\subsection{Experimental setup}
We equip a DeltaHand~\cite{10380685} with four tactile fingertips. All strain gauge sensors are calibrated for force estimation, as described in Sec.~\ref{sec:force-characterization}. The DeltaHand is mounted on a Franka Emika Panda robot arm. Force and vibrotactile sensing signals and arm and hand control signals are communicated through ROS with synchronization. We use Cartesian position control for both the Franka arm and the DeltaHand.

\subsection{Adaptive Grasping with Slip Detection}

We present a hand control policy combining force control and vibrotactile-based slip detection for adaptive grasping across diverse objects. The pipeline has two stages. 1) \emph{Pre-grasp}: each finger closes until the measured force \(F(t)\) reaches a threshold \(F_{\mathrm{th}} = 0.15\,\mathrm{N}\), ensuring a minimum contact force. 2) \emph{Lift with grasp adaptation}: the end-effector lifts the object while a \(30\,\mathrm{Hz}\) vibrotactile slip detector monitors contact stability. When a finger's vibrotactile amplitude exceeds the threshold \(0.006\), the finger closes by \(0.5\,\mathrm{mm}\) to increase grasp force. We evaluate success by whether the object is lifted without falling. To verify near-minimum-force grasping, after each successful lift, we gradually release the object with the same \(0.5\,\mathrm{mm}\) step size and record the force at slip onset as the minimum grasping force.


Fig.~\ref{fig:pinching-task}(a) shows the objects used for evaluation, including a fragile item (a potato chip piece), deformable items (a knitted hat, pencil bags, and a plush toy), as well as rigid and soft boxes and bags. Fig.~\ref{fig:pinching-task}(b) illustrates the grasping force $F_{test}$ achieved by using our method and the ground-truth minimum grasping force $F_{min}$ for each object. Each object is grasped five times and the average normal grasping force and the corresponding minimum grasping force are only computed over successful trials. The overall success rate is \(96\%\). The only failures occur for the knitted hat and the rigid USB hub. The thin and multilayer structure of the hat undergoes severe deformation when only the outer layer is pinched, causing the remaining layers to slip downward. The USB hub has smooth, rigid surfaces with low friction, which increases the likelihood of slippage when grasping the side faces. Successful grasps are achieved within \(0.3\,\mathrm{N}\) of the minimum grasp force for all objects, except a larger force gap \(0.6\,\mathrm{N}\) for the sand bag. This is because their particle-filled contents introduce additional vibrotactile disturbances, leading to false positive slip detections and higher grasping forces.

\subsection{Cup Localization and Unstacking}

\begin{figure*}[!ht]
\centering
\includegraphics[width=1\linewidth]{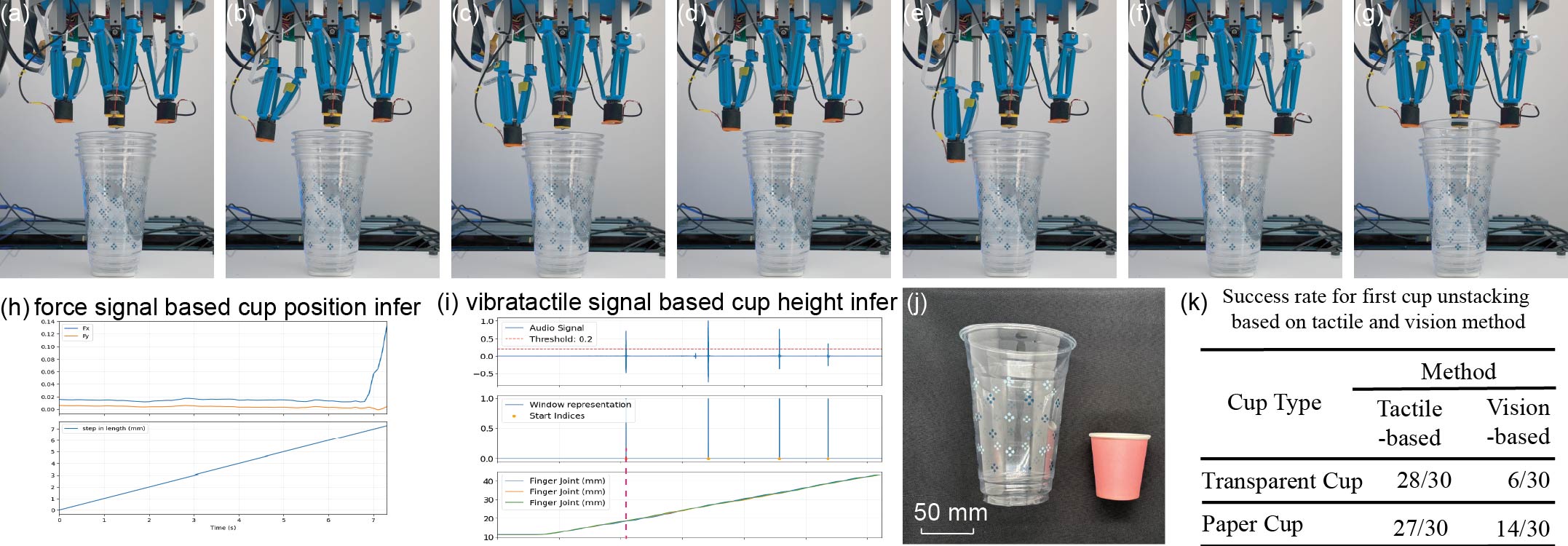}
\caption{Cup localization and unstacking. (a) The hand aligns with the cup stack; (b) A finger extends vertically; (c) It steps in horizontally to initiate the contact until a force threshold reached; (d) The finger resets above the stack; (e) It slides vertically along the stack edges; (f) Detect cup edges from vibrotactile features, guiding the fingers to the first edge height, and close until a force threshold (\(0.5~\mathrm{N}\)) is reached; (g) Separate and lift the top cup. (h) Cup edge localization using step-in distance and force readings. (i) Raw vibrotactile signals are binarized and synchronized with the finger positions to detect cup-edge contacts. (j) Visualization of transparent and paper cups. (k) Evaluation results for the tactile-based and vision-based methods.}

\label{fig:cups}
\vspace{-3mm}
\end{figure*}


We demonstrate our tactile fingertips for perceiving and manipulating fine-grained features in a cup unstacking task (Fig.~\ref{fig:cups}). We use force detection to locate the edge of the cups during horizontal approach (Fig.~\ref{fig:cups} (b) (c)) and vibrotactile signals to count the number of cup edges encountered during vertical sliding along the stack. By aligning vibrotactile features with the finger joint states (Fig.~\ref{fig:cups} (i)), we estimate the height of the top cup, and unstack it (Fig.~\ref{fig:cups}(f) (g)). 

As shown in Fig.~\ref{fig:cups} (i), the vibrotactile signals reveal distinct vibration patterns associated with each cup edge. To extract these vibrotactile features, we apply a threshold \(\tau\) to the time-domain signals to generate a binary sequence \(B(t)=\mathbf{1}\!\left[I(t)\ge \tau\right]\). A temporal clustering filter then merges segments separated by small gaps caused by noise. The onset of each resulting segment corresponds to contact between the fingernail and a cup edge. By synchronizing these events with the finger’s motion along the \(z\)-axis, we determine the corresponding cup edge's height. To unstack the top cup, the fingers are commanded to move to the target height and then close until reaching a predefined force threshold (\(0.5\,\mathrm{N}\)). To overcome inter-cup friction, two opposing fingers alternately move vertically to separate and lift the cup.


We test with two types of cups, plastic transparent cups and paper cups (Fig.~\ref{fig:cups} (j)), and conduct 30 trials for each. In each trial, the hand position was fixed, while the total number of cups in the stack was randomized to ${2, 3}$, or ${4}$. As a vision-based baseline to compare, we use a RealSense D415 camera placed in front of the workspace, and use YOLO26n-seg from the Ultralytics YOLO framework~\cite{jocher2023ultralytics} to segment the cups. The results are shown in  Fig.~\ref{fig:cups} (k), the vibratactile-based method achieved ($28/30$) success on transparent cups and ($27/30$) success on paper cups.  with failures mainly due to underestimating the distance to the cup stack and missing the edge catch during sliding. The vision baseline achieves \(6/30\) on transparent cups and \(14/30\) on paper cups. Failures are primarily due to segmentation errors with visual occlusion when the fingertip is in contact with the cups. 
These results show that multi-modal tactile sensing provides a stable and robust means of localizing fine-grained features and can operate effectively without vision.

\subsection{Hidden Object Perception through Active Interaction to Guide Manipulation}

\begin{figure*}[!h]
\centering
\includegraphics[width=1\linewidth]{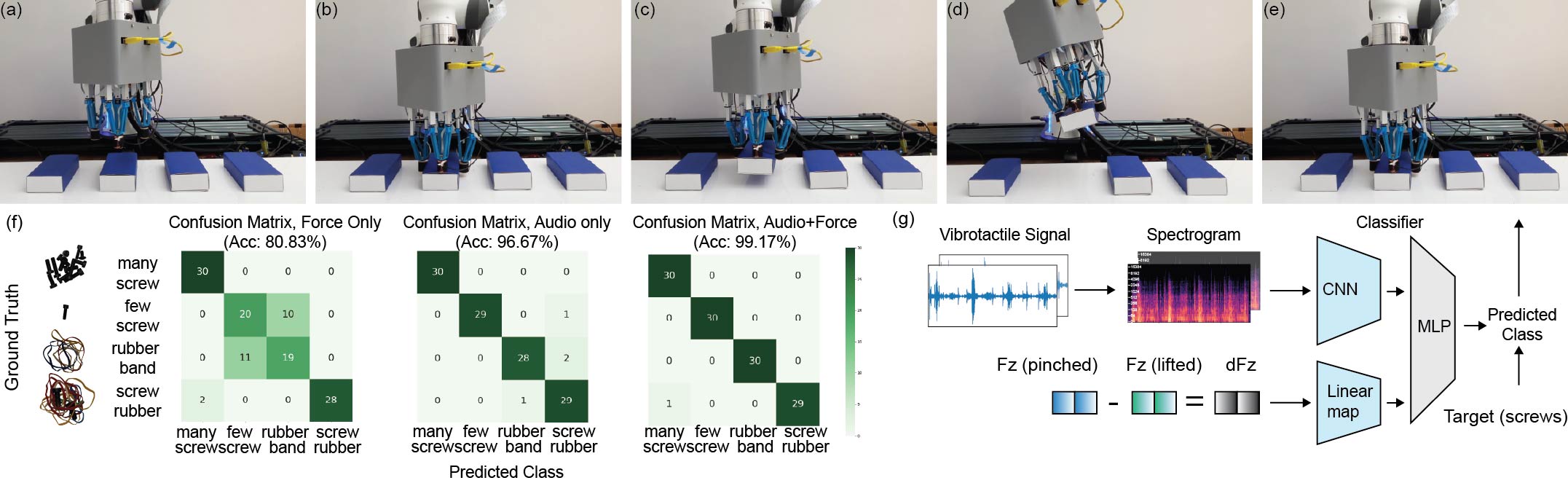}
\caption{Box selection and opening with occluded in-box object material identification. \textbf{Setup:} (a) Four visually identical boxes contain many screws, few screws, rubber bands, and a screw--rubber band mixture, representing contents with different stiffness and weight. \textbf{Pipeline:} (b) The robot grasps each box and records \(F_Z(\mathrm{pinched})\). (c) The robot lifts the box, holds it for 2~s, and records \(F_Z(\mathrm{lifted})\). (d) The robot shakes the box and records vibrotactile signals. (e) Based on the predicted class, the robot selects and opens the target box. \textbf{Results:} (f) Confusion matrices for force-only, vibrotactile-only, and fused-input classification. (g) The classifier predicts the hidden material class from shaking vibrotactile signals and the pre/post-lift force difference.}
\label{fig:shaking-setup}
\vspace{-3mm}
\end{figure*}

We investigate how to perceive hidden object properties with multi-modal tactile sensing through active interaction and use them to inform downstream manipulation with a box selection and opening task (Fig.~\ref{fig:shaking-setup}). Given four visually identical boxes, containing different materials (a few rubber bands, few screws, a mixture of rubber band and screws, and many screws), we explore each box with lifting and shaking motions, identify the material inside using a material classifier from tactile readings, and then select the box with the desired material to open.

During data collection, the robot is commanded to sequentially grasps, lifts, and shakes each box while recording force and vibrotactile signals. We use two opposing fingers to pinch the box to maintain the stable grasp during shaking. After lifting, the arm waits for \(2~\mathrm{s}\) to allow the force signal to stabilize, and then the z-axis force of both fingers is recorded. The robot then shakes the box by executing a pendulum-like rotation around the end effector in a vertical plane, with a maximum swing angle of \(30^\circ\) at \(0.5~\mathrm{Hz}\) for \(5~\mathrm{s}\), during which vibrotactile signals are recorded. We collected 140 recordings (35 per box) in total for material classifier training.

Raw vibrotactile signals are processed into spectrograms using a short-time Fourier transform (STFT), and embeddings (dim=2048) are extracted with a pre-trained CNN encoder (PANNs~\cite{kong2020panns}). The force signal is projected to the same embedding dimension of vibratactile signal via a linear layer. The concatenated force and vibrotactile embeddings are then fed into an MLP classifier with a 4D one-hot output corresponding to the four material classes.


During testing, the same lifting and shaking procedure is applied to a set of boxes. For each grasped box, the trained classifier performs a single forward pass to predict the in-box material class. Given the predicted label and a user-specified target material (Fig.~\ref{fig:shaking-setup}(g)), the robot selects the corresponding box and opens it by grasping on the side with two opposing fingers while using a third finger to slide and push the inner tray outward (Fig.~\ref{fig:shaking-setup}(e)).

As an ablation study, we choose different input modalities to the MLP classifier, including force-only, vibrotactile-only, and fused force–vibrotactile inputs. For each class, 30 evaluation trials are performed. The results are shown in Fig.~\ref{fig:shaking-setup}(f). Fused force and vibrotactile features achieves a 99.17\% success rate. In contrast, force-only input yields a 80.83\% success rate, with failures occurring between the few-rubber and few-screw classes due to their similar weights. Vibrotactile-only input also achieves 96.67\%, with misclassifications primarily between few and many rubber bands, as their compliant nature produces weak and less distinguishable vibratactile signal during shaking. These results demonstrate that multi-modal tactile sensing provides improved object physical property perception with complete visual occlusion compared to single modality, which can be further used to inform manipulation.

\section{Conclusions}
In this work, we present a low-cost, compact, multi-modal tactile fingertip for robotic hands. By integrating strain gauge sensors and a contact microphone sensor, the fingertip provides the sense of both static and dynamic contact information, i.e. forces and vibrations. Our characterization results show an estimate of 3D forces in the range of \(0\text{--}5~\mathrm{N}\) with \(0.07~\mathrm{N}\) MAE based on strain gauge sensors and an accuracy of \(95.49\%\) in material classification with sliding based on the contact microphone sensor. To demonstrate the functionality of the tactile fingertip, we equipped a DeltaHand with four proposed fingertips and evaluated with three manipulation tasks: (i) we achieved \(96\%\) success for tactile-based grasping with a wide range of objects using minimal forces; (ii) We demonstrated accurate cup edge localization and precise cup separation; and (iii) we showed how to identify hidden object properties through active exploration including lifting and shaking, and use it to inform downstream task i.e. box selection and open. These results demonstrate the effectiveness of using multi-modal tactile sensing in both low-level control loop as well as providing high-level environmental contexts through tactile-based perception with active interactions. In the future, we will further investigate how to unify the sensor fusion and skill learning with multi-modal tactile sensing to achieve more generalized manipulation with robotic hands.


\ifCLASSOPTIONcaptionsoff
  \newpage
\fi

\bibliographystyle{IEEEtran}
\bibliography{bibtex/bib/IEEEexample}

\end{document}